# Fair Kernel Learning

Adrián Pérez-Suay, Valero Laparra, Gonzalo Mateo-García, Jordi Muñoz-Marí,
Luis Gómez-Chova, and Gustau Camps-Valls [*]

Image Processing Laboratory (IPL), Universitat de València, Spain
C/ Cat. José Beltrán, 2. 46980 Paterna, València. Spain.
Adrian.Perez@uv.es, http://isp.uv.es

**Abstract.** New social and economic activities massively exploit big data and machine learning algorithms to do inference on people's lives. Applications include automatic curricula evaluation, wage determination, and risk assessment for credits and loans. Recently, many governments and institutions have raised concerns about the lack of fairness, equity and ethics in machine learning to treat these problems. It has been shown that not including sensitive features that bias fairness, such as gender or race, is not enough to mitigate the discrimination when other related features are included. Instead, including fairness in the objective function has been shown to be more efficient.

We present novel fair regression and dimensionality reduction methods built on a previously proposed fair classification framework. Both methods rely on using the Hilbert Schmidt independence criterion as the fairness term. Unlike previous approaches, this allows us to simplify the problem and to use multiple sensitive variables simultaneously. Replacing the linear formulation by kernel functions allows the methods to deal with nonlinear problems. For both linear and nonlinear formulations the solution reduces to solving simple matrix inversions or generalized eigenvalue problems. This simplifies the evaluation of the solutions for different trade-off values between the predictive error and fairness terms. We illustrate the usefulness of the proposed methods in toy examples, and evaluate their performance on real world datasets to predict income using gender and/or race discrimination as sensitive variables, and contraceptive method prediction under demographic and socio-economic sensitive descriptors.

## 1 Introduction

*"Perfect objectivity is an unrealistic goal; fairness, however, is not."* –M. Pollan, 2004

Current and upcoming application of machine learning to real-life's problems is overwhelming. Applications have enormous consequences in people's life, and impact decisions on education, economy, health care, and climate policies. The issue is certainly relevant. New social and economic activities massively exploit big data and machine learning algorithms to do inferences, and they decide on the best curriculum to fill in a position [15], to determine wages and in pre-trial risk assessment [4,9], and to evaluate risk of violence [8]. Companies, governments and institutions have raised concerns about the lack of fairness, equity and ethics in machine learning to treat this kind of problems[1]. Machine learning methods are actually far from being fair, just, or equitable in any way. After all, standard pattern analysis is often about model fitting and not the gender issue. Undoubtedly, attaining fair machine learning algorithms is a timely important concern. Fairness is an elusive concept though, so it is the inclusion of such *qualitative* measure in machines that only learn from data.

---

[*] The research was funded by the Spanish Ministry of Economy and Competitiveness (MINECO) through the projects TIN2015-64210-R and TEC2016-77741-R (ERDF).

Several approaches exist in the literature to account for fairness in machine learning. One of the earliest approaches tackled the bias problem through the definition of classification rules [23,25]. Later, some other works focused on (mainly) pre-processing the data [12,16,22,24]: down-weighting sensitive features or directly removing them have been the preferred choices. Perhaps the most naive approach is to simply discard the *sensitive* input features that bias discrimination [32]. Removing gender, disability or race, to predict monthly income is, however, not a good choice because model's accuracy may be largely impacted by the lack of informative features, and because some other correlated features enter the model anyway. This effect is known in statistics as the *omitted variable bias* [6].

Another simple approach consists on including *ad hoc* weights and data normalization to match the prior belief about fairness. Noting that data pre-processing is a quite arbitrary approach, Kamiran et al. proposed three solutions to learn fair classifiers [18]. Classifiers basically used the sensitive features only during learning and not at the prediction time. A step forward in this direction was presented in [12], where authors proposed pre-processing the data by removing information from all attributes correlated with the objective variable. The intuition behind this approach is that training on discrimination-free data is likely to yield more equitable predictions. A discussion of several more algorithms for binary protected and outcome variables can be found in [19]. Other authors have focused on finding transformations of the input space in order to extract features that do not retain information about the sensitive input variables [31].

All in all, the relevance of fair methods in machine learning is ever increasing, and a wide body of literature and approaches exist. We focus in this paper in a field known as 'disparate impact', in which outcomes should not differ based on individuals' protected class membership. Many definitions for the elusive concept of fairness in machine learning are available (see [3,5,7,11,12,17,23,30]): redlining, negative legacy, underestimation or subset targeting, to name a few. We frame our methods in the 'indirect discrimination' subfield.

Recently, an interesting regularization framework for fair classification was proposed in [20]. The framework optimizes a functional that jointly minimizes the classification error and the dependence between predictions and the sensitive variables using mutual-information concepts. We build our proposal upon this framework, and extend it to regression, and to unsupervised dimensionality reduction problems with kernel methods. The proposed kernel machines exploit cross-covariance operators in Hilbert spaces. Both theoretical and empirical advantages are gained. Advantageously, the solutions only involve solving simple matrix inversion or generalized eigenproblems. This allows to check different solutions when the trade-off between prediction and fairness is modified. The methods are able to deal naturally with input variables of several dimensions for the regular as well as for the sensitive variables. Note that this is especially important for the fairness term, where a robust measure of dependence is needed. On top of this, the proposed methods can incorporate prior knowledge about the fairness, invariances and interestingness of the feature relations. We illustrate performance in toy data as well as in two real problems: income prediction subject to gender and/or race discrimination, and contraceptive method prediction under demographic and socio-economic sensitive descriptors.

---

[1] http://www.fatml.org/

The remainder of the paper is organized as follows. Section 2 describes the problem statement, introduces notation and presents the fair kernel regression framework in the input and the Hilbert space. Section 3 extends the fair kernel learning framework to dimensionality reduction problems. Toy examples guide the presentation of the two approaches. Experimental evidence of performance is given in Section 4. Conclusions finalize the paper in Section 5.

## 2 Fair Regression Framework

This section starts by defining the notation and the concept of fair predictions. Then we introduce the proposed framework for performing fair regression learning based on cross-covariance operators for dependence estimation in Hilbert spaces. We conclude with an illustrative example.

### 2.1 Notation, preliminaries, and the regularization framework

Let us define the notation and the problem setting. We are given $n$ samples of a response (or target) data matrix $\mathbf{Y} \in \mathbb{R}^{n \times c}$, and $d + q$ prediction variables: $d$ unprotected $\mathbf{X}_u \in \mathbb{R}^{n \times d}$ and $q$ sensitive $\mathbf{S} \in \mathbb{R}^{n \times q}$. The goal is to obtain an accurate prediction function (or model) $f$ for the target variable $\mathbf{Y}$ from the input data, $\mathbf{X} = (\mathbf{X}_u, \mathbf{S})$. This function is said to be *totally fair* if the predictions are statistically independent of the sensitive (protected) features [5,10].

Therefore, two main ingredients are needed to perform fair predictions: we need to ensure independence of the predictions on the sensitive variables, and simultaneously to obtain a good approximation of the target variables. The regularization framework proposed in [20] tackles the problem of finding a fair function $f$ for classification by including a term to enforce fair classification. In our proposal the proposed function $f$ tries to learn the relation between observed input-output data pairs $(\mathbf{x}_1, \mathbf{y}_1), \ldots, (\mathbf{x}_n, \mathbf{y}_n) \in \mathcal{X} \times \mathcal{Y}$ such that generalizes well

(good predictions $\hat{\mathbf{y}}_* = f(\mathbf{x}_*) \in \mathcal{Y}$ for the unseen input data point $\mathbf{x}_* \in \mathcal{X}$), and the predictions should be as independent as possible of the sensitive features. Then, the following functional needs to be optimized:

$$\mathcal{L} = \frac{1}{n}\sum_{i=1}^{n} V(f(\mathbf{x}_i), \mathbf{y}_i) + \lambda \ \Omega(\|f\|_{\mathcal{H}}) + \mu \ I(f(\mathbf{x}), \mathbf{s}), \tag{1}$$

where $V$ is the error cost function, $\Omega(\|f\|_{\mathcal{H}})$ acts as a regularizer of the predictive function and controls the smoothness and complexity of the model, and $I(f(\mathbf{x}), \mathbf{s})$ measures the independence between model's predictions and the protected variables. Note that one aims to minimize the amount of information that the model shares with the sensitive variables while controlling the trade-off between fitting and independence through hyperparameters $\lambda$ and $\mu$. By setting $\mu = 0$ one obtains the ordinary Tikhonov's regularized functional, and by setting $\lambda = 0$ one obtains the unregularized versions of this framework.

The framework admits many variants depending of the cost function $V$, regularizer $\Omega$ and the independence measure, $I$. For example, in [20], the function $f$ was the logistic regression classifier and $I$ was a simplification of the mutual information estimate. Despite the good results reported in [20], these choices did not allow to solve the problem in closed-form, nor coping with more than one sensitive variable at the same time, since the proposed mutual information is an uni-dimensional dependence measure. In the following section, we elaborate further this framework under the concept of cross-covariance operators in Hilbert spaces, which lead to closed-form solutions and allow to deal with several sensitive variables simultaneously.

### 2.2 Fair Linear Regression

Let us now provide a straightforward instantiation of the proposed framework for fair linear regression (FLR). We will adopt a linear predictive model for $f$, i.e. the matrix of predictions for a test data matrix $\mathbf{X}_*$ is given by $\hat{\mathbf{Y}}_* = \mathbf{X}_*\mathbf{W}$, the mean square error for the cost function $V = \|\mathbf{Y} - \mathbf{XW}\|_2^2$ and the standard $\ell_2$ regularization for model weights $\Omega := \|\mathbf{W}\|_2^2$. Other choices could be taken, leading to alternative formulations. In order to measure dependence, we will rely on the cross-covariance operator between the predictions and the sensitive variables in Hilbert space. Let us consider two spaces $\mathcal{Y} \subseteq \mathbb{R}^c$ and $\mathcal{S} \subseteq \mathbb{R}^q$, where random variables $(\hat{\mathbf{y}}, \mathbf{s})$ are sampled from the joint distribution $\mathbb{P}_{\mathbf{ys}}$. Given a set of pairs $\mathcal{D} = \{(\hat{\mathbf{y}}_1, \mathbf{s}_1), \ldots, (\hat{\mathbf{y}}_n, \mathbf{s}_n)\}$ of size $n$ drawn from $\mathbb{P}_{\mathbf{ys}}$, an empirical estimator of HSIC [14] allows us to define

$$I := \mathrm{HSIC}(\mathcal{Y}, \mathcal{S}, \mathbb{P}_{\mathbf{ys}}) = \|\mathbf{C}_{ys}\|_{\mathrm{HS}}^2 = \|\tilde{\mathbf{Y}}^\top \tilde{\mathbf{S}}\|^2 = \frac{1}{n^2}\mathrm{Tr}(\tilde{\mathbf{Y}}^\top \tilde{\mathbf{S}} \tilde{\mathbf{S}}^\top \tilde{\mathbf{Y}}),$$

where $\|\cdot\|_{\mathrm{HS}}$ is the Hilbert-Schmidt norm, $\mathbf{C}_{ys}$ is the empirical cross-covariance matrix between predictions and sensitive variables[2], $\tilde{\mathbf{Y}}$ and $\tilde{\mathbf{S}}$ represent the feature-centered $\mathbf{Y}$ and $\mathbf{S}$ respectively, and $\mathrm{Tr}(\cdot)$ denotes the trace of the matrix. We want to stress that HSIC allows us to estimate dependencies between multi-dimensional variables, and that HSIC is zero if an only if there is no second-order dependence between $\hat{\mathbf{y}}$ and $\mathbf{s}$. In the next section we extend the formulation to higher-order dependencies with the use of kernels [26,27].

---

[2] The covariance matrix is $\mathcal{C}_{\mathbf{ys}} = \mathbb{E}_{\mathbf{ys}}(\mathbf{ys}^\top) - \mathbb{E}_{\mathbf{y}}(\mathbf{y})\mathbb{E}_{\mathbf{s}}(\mathbf{s}^\top)$, where $\mathbb{E}_{\mathbf{ys}}$ is the expectation with respect to $\mathbb{P}_{\mathbf{ys}}$, and $\mathbb{E}_{\mathbf{y}}$ is the marginal expectation with respect to $\mathbb{P}_{\mathbf{y}}$ (hereafter we assume that all these quantities exist).

Plugging these definitions of $f$, $V$, $\Omega$ and $I$ in Eq. (1), one can easily show that the solution has the following closed-form solution for weight estimates

$$\widehat{\mathbf{W}} = (\tilde{\mathbf{X}}^\top \tilde{\mathbf{X}} + \lambda\,\mathbf{I} + \frac{\mu}{n^2}\,\tilde{\mathbf{X}}^\top \tilde{\mathbf{S}} \tilde{\mathbf{S}}^\top \tilde{\mathbf{X}})^{-1} \tilde{\mathbf{X}}^\top \mathbf{Y}, \qquad (2)$$

where fairness is trivially controlled with $\mu$, which acts as an additional regularization term. Also note that when $\mu = 0$ the ordinary regularized least squares solution is obtained.

### 2.3 Fair Kernel Regression

Let us now extend the previous model to the nonlinear case in terms of the prediction function, the regularizer and the dependence measure by means of reproducing kernels [26,27]. We call this method the fair kernel regression (FKR) model. We proceed in the standard way in kernel machines by mapping data $\mathbf{X}$ and $\mathbf{S}$ to a Hilbert space $\mathcal{H}$ via the mapping functions $\phi(\cdot)$ and $\psi(\cdot)$ respectively. This yields $\boldsymbol{\Phi}, \boldsymbol{\Psi} \in \mathcal{H} \subseteq \mathbb{R}^{d_\mathcal{H}}$, where $d_\mathcal{H}$ is the (unknown and possibly infinite) dimensionality of mapped points in $\mathcal{H}$. The corresponding kernel matrices can be defined as: $\tilde{\mathbf{K}} = \tilde{\boldsymbol{\Phi}} \tilde{\boldsymbol{\Phi}}^\top$ and $\tilde{\mathbf{K}}_S = \tilde{\boldsymbol{\Psi}} \tilde{\boldsymbol{\Psi}}^\top$. Now the prediction function is $\hat{\mathbf{Y}} = \boldsymbol{\Phi} \mathbf{W}_\mathcal{H}$, the regularizer is $\Omega := \|\mathbf{W}_\mathcal{H}\|_2^2$, and the dependence measure $I$ is the HSIC estimate between predictions $\hat{\mathbf{Y}}$ and sensitive variables $\mathbf{S}$, which can now be estimated in Hilbert spaces: $I := \mathrm{HSIC}(\mathcal{Y}, \mathcal{H}, \mathbb{P}_{\mathbf{ys}}) = \|\mathbf{C}_{ys}\|_{\mathrm{HS}}^2$. Now, by plugging all these terms in the cost function, using the representer's theorem $\mathbf{W}_\mathcal{H} = \tilde{\boldsymbol{\Phi}}^\top \boldsymbol{\Lambda}$ and after some simple linear algebra, we obtain the dual weights in closed-form

$$\boldsymbol{\Lambda} = (\tilde{\mathbf{K}} + \lambda \mathbf{I} + \frac{\mu}{n^2} \tilde{\mathbf{K}} \tilde{\mathbf{K}}_S)^{-1} \mathbf{Y}, \qquad (3)$$

which can be used for prediction with a new point $\mathbf{x}_*$ by using $\hat{\mathbf{y}}_* = \mathbf{k}_* \boldsymbol{\Lambda}$, where $\mathbf{k}_* = [K(\mathbf{x}_*, \mathbf{x}_1), \ldots, K(\mathbf{x}_*, \mathbf{x}_n)]^\top$. In the case where $\mu = 0$ the method reduces to standard kernel ridge regression (KRR) method [27]. Centering points in feature spaces can be done implicitly with kernels [27]: a kernel matrix $\mathbf{K}$ is centered by doing $\tilde{\mathbf{K}} = \mathbf{H} \mathbf{K} \mathbf{H}$, where $\mathbf{H} = \mathbf{I} - \frac{1}{n} \mathbb{1} \mathbb{1}^\top$.

**Lemma 1.** *Both KRR and FKR model weights are bounded in norm by the same quantitiy.*

*Proof.* Let us assume the same kernel matrix $\tilde{\mathbf{K}}$ for KRR and FKR, and also suppose $\lambda, \mu \geq 0$, then the following bound is satisfied: $\|(\tilde{\mathbf{K}} + \lambda \mathbf{I} + \frac{\mu}{n^2} \tilde{\mathbf{K}}_S \tilde{\mathbf{K}})^{-1}\| \leq \|(\tilde{\mathbf{K}} + \lambda \mathbf{I})^{-1}\|$. Given $\mu \geq 0$ the following inequality, with $\succeq$ meaning the standard PSD order, holds true: $\tilde{\mathbf{K}} + \lambda \mathbf{I} + \frac{\mu}{n^2} \tilde{\mathbf{K}}_S \tilde{\mathbf{K}} \succeq \tilde{\mathbf{K}} + \lambda \mathbf{I}$. Then also holds $(\tilde{\mathbf{K}} + \lambda \mathbf{I} +$

$\frac{\mu}{n^2}\tilde{\mathbf{K}}_S\tilde{\mathbf{K}})^{-1} \preceq (\tilde{\mathbf{K}}+\lambda\mathbf{I})^{-1}$, and by taking norms we have the following inequality $\|(\tilde{\mathbf{K}}+\lambda\mathbf{I}+\frac{\mu}{n^2}\tilde{\mathbf{K}}_S\tilde{\mathbf{K}})^{-1}\| \leq \|(\tilde{\mathbf{K}}+\lambda\mathbf{I})^{-1}\|$. FKR model weights can be bounded

$$\begin{aligned}\|\boldsymbol{\Lambda}_{\text{FKR}}\| &= \left\|(\tilde{\mathbf{K}}+\lambda\mathbf{I}+\frac{\mu}{n^2}\tilde{\mathbf{K}}_S\tilde{\mathbf{K}})^{-1}\mathbf{Y}\right\| \leq \left\|(\tilde{\mathbf{K}}+\lambda\mathbf{I}+\frac{\mu}{n^2}\tilde{\mathbf{K}}_S\tilde{\mathbf{K}})^{-1}\right\|\|\mathbf{Y}\|\\ &\leq \left\|(\tilde{\mathbf{K}}+\lambda\mathbf{I})^{-1}\right\|\|\mathbf{Y}\|,\end{aligned} \quad (4)$$

which is the same bound for KRR weights:

$$\|\boldsymbol{\Lambda}_{\text{KRR}}\| = \left\|(\tilde{\mathbf{K}}+\lambda\mathbf{I})^{-1}\mathbf{Y}\right\| \leq \left\|(\tilde{\mathbf{K}}+\lambda\mathbf{I})^{-1}\right\|\|\mathbf{Y}\|.$$

*Illustrative example.* Here we illustrate the performance of the proposed methods in a controlled synthetic experiment. The data considers a sensitive variable drawn from a zero mean Gaussian with standard deviation $\sigma_s$, $\mathbf{s} \sim \mathcal{N}(0, \sigma_s)$, and a parametric function $f_s(\mathbf{s})$ that yields an intermediate variable $\mathbf{a}$ buried in additive white Gaussian noise (AWGN), i.e. $\mathbf{a} = f_s(\mathbf{s})+\mathbf{n}_f$, where $\mathbf{n}_f \sim \mathcal{N}(0, \sigma_f)$. System's output combines both the sensitive as well as its arbitrarily transformed version affected by AWGN $\mathbf{y} = g_s(\mathbf{s}) + g_r(\mathbf{a}) + \mathbf{n}_y$, where $\mathbf{n}_y \sim \mathcal{N}(0, \sigma_y)$. In this example we used $f_s(x) = \log(x)$, and $g_r(x) = g_s(x) = x^2$. This system ensures that, even without using variable $\mathbf{s}$ explicitly as an input factor in the regression model, the information conveyed in $\mathbf{s}$ is embedded in $\mathbf{a}$ indirectly. Two settings are considered, with and without using the sensitive variable $\mathbf{s}$ as an input feature. In both experiments we used the RBF kernel function and fitted hyperparameters ($\lambda$, $\mu$ and the kernel widths) to be optimal for each $\mu$ value. Figure 1 shows the results for the four different configurations (linear and nonlinear, with and without considering $\mathbf{s}$), the horizontal axis represents the mean square error (MSE) of the prediction and the vertical axis the HSIC between the prediction and the protected variable. An ideal fair model would obtain zero MSE and zero HSIC. For each configuration, we give the family of solutions that can be obtained by modifying the parameter $\mu$. Classical solutions that do not include the fairness term show that KRR improves the ordinary LR results in MSE terms, but both methods obtain similar HSIC values. On the other hand, the inclusion of the sensitive variable $\mathbf{s}$ as input feature obtains more fair results in HSIC terms but worst results in MSE terms. The fairness paths are obtained for different $\mu$ values. The nonlinear regression methods outperform in general the linear counterparts. Including the sensitive variable as input returns better trade-off results. For example, the FKR can be tuned to have the same fairness level as KRR\S but obtaining around 30% lower prediction error. A similar conclusion can be extracted in the linear case, yet the improvement is smaller.

## 3 Fair Dimensionality Reduction Framework

Let us now show a different frame for fair machine learning. Rather than optimizing a regression model, we are here concerned about obtaining fair feature representations. We rely on the field of multivariate analysis to develop both linear and nonlinear (kernel) dimensionality reduction methods.

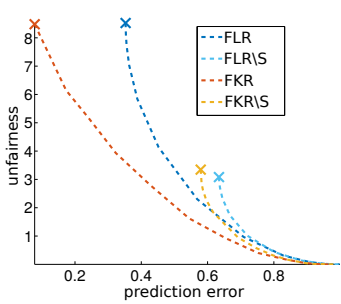

**Fig. 1.** Regression curves of the prediction error (MSE) versus the unfairness (independence of predictions with sensitive variables) in four different configurations (FLR and FKR, with and without the sensitive variable **s**) and different values of $\mu$ (crosses indicate $\mu = 0$ solutions).

### 3.1 Fair Dimensionality Reduction

Let us define two training data matrices as before, the full design matrix $\mathbf{X} \in \mathbb{R}^{n \times d}$, and the *sensitive* data matrix $\mathbf{S} \in \mathbb{R}^{n \times q}$, and a labelled data matrix $\mathbf{Y} \in \mathbb{R}^{n \times c}$ (here we use 1-of-$c$ encoding). The goal here is to find a *fair projection matrix* $\mathbf{V} \in \mathbb{R}^{d \times n_p}$ such that the projected data, $\mathbf{XV}$ keeps as much information as possible from the input features yet minimally aligned with the protected, sensitive features. We denoted $\mathbf{V} = [\mathbf{v}_1|\cdots|\mathbf{v}_{n_p}]$, where $\mathbf{v}_i$ is the $i$-th projection vector and $n_p$ is the dimension of the projection subspace. Hereafter, the terms alignment and statistical dependency will be used interchangeably. As before, in order to minimize alignment (dependence) between random variables $\mathbf{X}$ and $\mathbf{S}$, we will use the cross-covariance operator, whose empirical estimate reduces to compute the norm of the corresponding empirical cross-covariance given by the HSIC estimator. We will also use HSIC to maximize the dependence between the projected and the original data. The problem can thus be easily formalized as the maximization of the following Rayleigh quotient:

$$\mathbf{V}^* = \arg\max_{\mathbf{V}} \left\{ \frac{\text{HSIC}(\tilde{\mathbf{X}}\mathbf{V}, \tilde{\mathbf{X}})}{\text{HSIC}(\tilde{\mathbf{X}}\mathbf{V}, \tilde{\mathbf{S}})} \right\} = \arg\max_{\mathbf{V}} \left\{ \frac{\text{Tr}(\mathbf{V}^\top(\tilde{\mathbf{X}}^\top\tilde{\mathbf{X}}\tilde{\mathbf{X}}^\top\tilde{\mathbf{X}})\mathbf{V})}{\text{Tr}(\mathbf{V}^\top(\tilde{\mathbf{X}}^\top\tilde{\mathbf{S}}\tilde{\mathbf{S}}^\top\tilde{\mathbf{X}})\mathbf{V})} \right\},$$

where $\tilde{\mathbf{X}}$ represents the feature-centered $\mathbf{X}$. This leads to solving a generalized eigenvalue problem with the empirical covariance $\mathbf{C}_{xx} = \frac{1}{n}\tilde{\mathbf{X}}^\top\tilde{\mathbf{X}}$ and input-sensitive cross-covariance $\mathbf{C}_{xs} = \frac{1}{n}\tilde{\mathbf{X}}^\top\tilde{\mathbf{S}}$:

$$\mathbf{C}_{xx}\mathbf{C}_{xx}^\top \mathbf{v} = \lambda \mathbf{C}_{xs}\mathbf{C}_{xs}^\top \mathbf{v}.$$

The solution resembles that of the standard orthonormalized partial least squares (OPLS) [28]. Note that the generalized eigenproblem involves symmetric matrices. The matrix projection operator $\mathbf{V}$ can then been used to obtain *fair scores* for a new data point $\mathbf{x}_* \in \mathbb{R}^{d \times 1}$ as follows $\tilde{\mathbf{x}}'_* = \mathbf{V}^\top \mathbf{x}_* \in \mathbb{R}^{n_p \times 1}$.

### 3.2 Kernel Fair Dimensionality Reduction

Let us now derive a nonlinear version of FDR by means of reproducing kernels [26, 27]. We proceed in the standard way by mapping data $\mathbf{X}$ and $\mathbf{S}$ to a Hilbert space $\mathcal{H}$ via mapping functions $\phi(\cdot), \psi(\cdot)$, which yield $\boldsymbol{\Phi}, \boldsymbol{\Psi} \in \mathbb{R}^{n \times d_\mathcal{H}}$ respectively, where $d_\mathcal{H}$ is the dimensionality of $\mathcal{H}$. The FDR ratio now translates into finding a projection matrix $\mathbf{U} = [\mathbf{u}_1|\cdots|\mathbf{u}_{n_p}] \in \mathbb{R}^{d_\mathcal{H} \times n_p}$ such that:

$$\mathbf{U}^* = \arg\max_{\mathbf{U}} \left\{ \frac{\mathrm{Tr}(\mathbf{U}^\top(\tilde{\boldsymbol{\Phi}}^\top\tilde{\boldsymbol{\Phi}}\tilde{\boldsymbol{\Phi}}^\top\tilde{\boldsymbol{\Phi}})\mathbf{U})}{\mathrm{Tr}(\mathbf{U}^\top(\tilde{\boldsymbol{\Phi}}^\top\tilde{\boldsymbol{\Psi}}\tilde{\boldsymbol{\Psi}}^\top\tilde{\boldsymbol{\Phi}})\mathbf{U})} \right\},$$

where $\tilde{\boldsymbol{\Phi}}$, and $\tilde{\boldsymbol{\Psi}}$ contain the *centered* data in Hilbert space. Now, by applying the representer's theorem $\mathbf{U} = \tilde{\boldsymbol{\Phi}}^\top \boldsymbol{\Lambda}$ (where $\boldsymbol{\Lambda} = [\boldsymbol{\alpha}_1|\cdots|\boldsymbol{\alpha}_{n_p}]^\top$), replacing dot products with kernel functions, $\tilde{k}_x(\mathbf{x}, \mathbf{x}') = \tilde{\phi}(\mathbf{x})^\top \tilde{\phi}(\mathbf{x}')$, $\tilde{k}_s(\mathbf{s}, \mathbf{s}') = \tilde{\psi}(\mathbf{s})^\top \tilde{\psi}(\mathbf{s}')$, and defining kernel matrices, $\tilde{\mathbf{K}}_x = \tilde{\boldsymbol{\Phi}}\tilde{\boldsymbol{\Phi}}^\top$, and $\tilde{\mathbf{K}}_s = \tilde{\boldsymbol{\Psi}}\tilde{\boldsymbol{\Psi}}^\top$, we obtain a dual problem:

$$\boldsymbol{\Lambda}^* = \arg\max_{\boldsymbol{\Lambda}} \left\{ \frac{\mathrm{Tr}(\boldsymbol{\Lambda}^\top(\tilde{\mathbf{K}}_x\tilde{\mathbf{K}}_x\tilde{\mathbf{K}}_x)\boldsymbol{\Lambda})}{\mathrm{Tr}(\boldsymbol{\Lambda}^\top(\tilde{\mathbf{K}}_x\tilde{\mathbf{K}}_s\tilde{\mathbf{K}}_x)\boldsymbol{\Lambda})} \right\},$$

which reduces again to solving a generalized eigenproblem:

$$\tilde{\mathbf{K}}_x\tilde{\mathbf{K}}_x\boldsymbol{\alpha} = \lambda\tilde{\mathbf{K}}_s\tilde{\mathbf{K}}_x\boldsymbol{\alpha}.$$

This problem can be solved iteratively by first computing the leading pair $\{\lambda_i, \boldsymbol{\alpha}_i\}$, and then deflating the matrices. The deflation equation for KFDR can be written as:

$$\tilde{\mathbf{K}}_x\tilde{\mathbf{K}}_x\tilde{\mathbf{K}}_x \leftarrow \tilde{\mathbf{K}}_x\tilde{\mathbf{K}}_x\tilde{\mathbf{K}}_x - \lambda_i\tilde{\mathbf{K}}_x\tilde{\mathbf{K}}_s\tilde{\mathbf{K}}_x\boldsymbol{\alpha}_i\boldsymbol{\alpha}_i^\top\tilde{\mathbf{K}}_x\tilde{\mathbf{K}}_s\tilde{\mathbf{K}}_x.$$

which is equivalent to

$$\tilde{\mathbf{K}}_x \leftarrow \tilde{\mathbf{K}}_x - \sqrt{\lambda_i}\tilde{\mathbf{K}}_x\tilde{\mathbf{K}}_s\boldsymbol{\alpha}_i,$$

i.e., at each step we remove from the kernel matrix the best approximation based on the newly extracted projections of the sensitive data $\tilde{\mathbf{K}}_x\tilde{\mathbf{K}}_s\boldsymbol{\alpha}_i$. The deflation procedure decreases by 1 the rank of the matrix, so the maximum number of features that can be extracted with KFDR is $\mathrm{rank}(\tilde{\mathbf{K}}_x\tilde{\mathbf{K}}_s)$, which for most mapping functions will be $n_p = \min\{n, c\}$.

The KFDR method is again similar to the KOPLS in [1, 2], but here we seek for independent projections from the inequitable variables $\mathbf{S}$ while maximizing the variance. As for any kernel multivariate analysis method, projecting a new test point $\mathbf{x}_* \in \mathbb{R}^{d \times 1}$ is possible, $\tilde{\mathbf{x}}'_* = \mathbf{U}^\top \tilde{\phi}(\mathbf{x}_*) = \boldsymbol{\Lambda}^\top \tilde{\boldsymbol{\Phi}}\tilde{\phi}(\mathbf{x}_*) = \boldsymbol{\Lambda}^\top \tilde{\mathbf{k}}_*$, where $\tilde{\mathbf{k}}_* = [k_x(\mathbf{x}_1, \mathbf{x}_*), \ldots, k_x(\mathbf{x}_n, \mathbf{x}_*)]^\top$.

*Invariant feature extraction.* This example considers $n = 1000$ points drawn from a sinus function buried in noise, $b_i = \sin(a_i) + n_i$, where $\mathbf{a} \sim \mathcal{U}(0, 1.5\pi)$ and $n_i \sim \mathcal{N}(0, 0.1)$. We compare the maps of PCA and FDR and their kernel counterparts. For illustration purposes we consider two different configurations of the inputs, by switching the *sensitive* variable to be either **a** or **b**. Note that this only changes the results for FDR since PCA and KPCA do not distinguish between *sensitive* and *unprotected* variables. Figure 2 shows the first component projection as a color map in the background for the different methods. Essentially PCA and FDR methods cannot account for the nonlinear feature relations, but FDR allows one to easily force invariance to a pre-specified dimension of interest. Compare for instance the first (PCA) and the second (FDR, **S** = **a**) plots. The first component found by PCA is diagonal, revealing it has information about both components (**a** and **b**). On the other hand, the first component found by FDR is vertical, thus avoiding the information in the horizontal axis, i.e. it is insensitive to the information in **a**, as expected. Similar effects are observed in the kernel versions, yet recovering the nonlinear structure of the manifold.

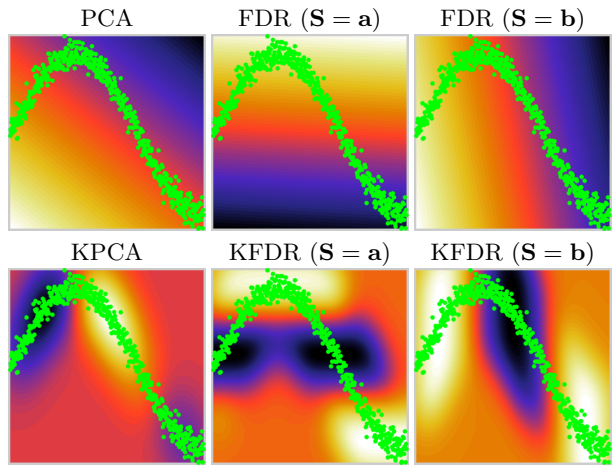

**Fig. 2.** Linear and kernel feature extraction in a noisy sinusoid distribution. For the sake of simplicity we only show projections onto the first component. The value of the projection is shown as a color map in the background, where dark tones mean small values and brighter tones mean big values. See details in the text.

*Noise-aware feature extraction.* We generated a bidimensional banana-shaped distribution corrupted by correlated noise in the $\pi/4$-direction to which we want to be independent, cf. Fig. 3. We compare results of KFDR with those from standard KPCA. Projections #2 and #3 capture the noise distribution while for the KFDR all extracted projections are invariant to variations in the $\pi/4$ direction where the noise is mostly present. The method is intimately related to the kernel signal to noise ratio in [13].

## 4   Experimental Evidence

The aim of this section is to empirically test the proposed methods on real data. We will see that using regular models and removing the sensitive variables is not enough to obtain fair solutions. First, we will present the data used and then we will evaluate both proposals, regression and dimensionality reduction, using two different datasets from the UCI Machine Learning repository. Source code and illustrative demos are available in http://isp.uv.es/soft_regression.html for the interested reader.

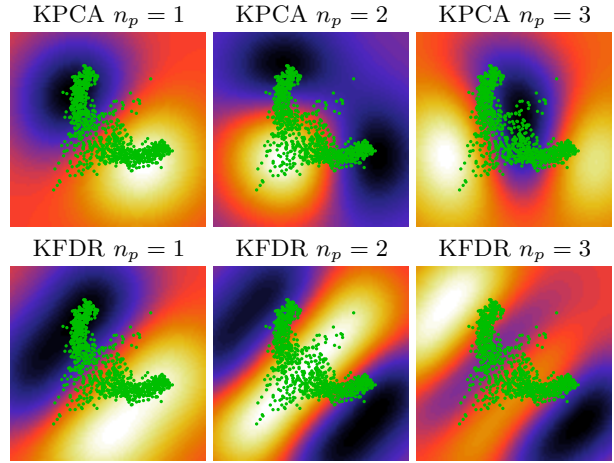

**Fig. 3.** Dimensionality reduction in a noisy two-dimensional example.

### 4.1 Datasets

We consider two datasets from the UCI repostory [21]: the Adult dataset and the Contraceptive dataset. Both of them involve sensitive attributes and pose problems of equitable prediction.

**Dataset 1: Adult dataset** This dataset is readily pre-processed and available from the libsvm website[3], and has been used in several studies about fair machine learning methods for classification and feature extraction [12, 20, 29, 31]. The original Adult data set contains 14 features, among which six are continuous and eight are categorical. In this data set, continuous features were discretized into quantiles, and each quantile was represented by a binary feature. Also, a categorical feature with $m$ categories is converted to $m$ binary features. Finally, the original 14 features are pre-processed into 123 features. Details on how each feature is finally converted can be found in Table 1. The dataset is already split into 2 sets, the first one for training the models, which consists of 32561 instances, and the second one was used for testing the results and contains 16281 instances. Both in regression and dimensionality reduction experiments we fit the hyperparameters using 5000 instances to train and 5000 to validate both randomly selected from the training set. Afterwards we evaluate those models using the whole test set. All the presented results are the mean of twenty-five realizations of each experiment.

---

[3] https://www.csie.ntu.edu.tw/~cjlin/libsvmtools/datasets/binary.html

**Table 1.** Original and processed features for the adult dataset from the UCI repository. The type column distinguishes between a continuous (c) or discrete (d) attribute.

| # feature | original feature | type (c/d) | # processed feat. |
|---|---|---|---|
| 1 | age | c | [1,5] |
| 2 | workclass | d | [6,13] |
| 3 | final weight | c | [14,18] |
| 4 | education | d | [19,34] |
| 5 | ed_num | c | [35,39] |
| 6 | marital_status | d | [40,46] |
| 7 | occupation | d | [47,60] |
| 8 | relationship | d | [61,66] |
| 9 | race | d | [67,71] |
| 10 | sex | d | [72,73] |
| 11 | capital_gain | c | [74,75] |
| 12 | capital_loss | c | [76,77] |
| 13 | hours × week | c | [78,82] |
| 14 | country | d | [83,123] |

**Dataset 2: Contraceptive Method Choice Data Set** In the second problem we study the drivers for adoption of contraceptive types by a women cohort. We used the Contraceptive Method Choice (CMC) Data Set from the UCI repository, which can be downloaded from https://archive.ics.uci.edu/ml/datasets/Contraceptive+Method+Choice. This dataset is a subset of the 1987 National Indonesia Contraceptive Prevalence Survey. The samples are married women who were either not pregnant or do not know if they were at the time of interview. The problem is to predict the current contraceptive method choice leading to three possibilities: 'no use', 'long-term methods', or 'short-term methods' of a woman based on demographic and socio-economic descriptors. We simplified the problem and considered the classes using/not-using a contraceptive method. Table 2 summarizes the total number of features and the class attributes.

**Table 2.** Original and processed features for the contraceptive method choice data set from the UCI repository. The type column distinguishes between a continuous (c) or discrete (d) attributes.

| # feature | original feature | type (c/d) |
|---|---|---|
| 1 | wife's age | c |
| 2 | wife's education | d |
| 3 | husband's education | d |
| 4 | number of children ever born | c |
| 5 | wife's religion | d |
| 6 | wife's now working | d |
| 7 | husband's occupation | d |
| 8 | standard-of-living index | d |
| 9 | media exposure | d |
| 10 | contraceptive method used | class attribute |

The data set consists of 1473 samples with 9 features, and one variable to infer, the contraceptive method. In order to train our algorithms, we split the data into train (500 samples), validation (500 samples) and test sets (the remaining 473 samples). The experiment is performed 25 times, and results are averaged to avoid skewed conclusions.

### 4.2 Experimental setup

In the regression experiment, we optimize the hyperparameters $\lambda$ (model regularization), $\sigma$ (kernel width) and $\sigma_S$ (the kernel parameter for the dependence estimation) using different logarithmically spaced values. Specifically we tried seven values in the interval $[10^{-4}, 10^3]$ for $\lambda$, 10 values in $[10^{-4}, 10^4]$ for $\sigma$, and 10 values in $[10^{-1}, 10^2]$ for $\sigma_S$. We start by seeking the optimal $\lambda$ and $\sigma$ parameters that minimize the error in the validation data. Once these two parameters are fixed we explore the kernel parameter for the dependence estimation in order to maximize the dependence between the model and the sensitive data. Finally, we try 25 different logarithmic spaced values in the interval $[10^{-7}, 10^3]$ for the $\mu$ fairness hyperparameter (large $\mu$ values imply more fair models).

In the FDR experiment the only hyperparameter to tune is $\sigma_S$, which is optimized to maximize the dependence between the transformed data and the sensitive variables. We optimized this parameter trying 15 values in the interval $[10^{-5}, 10^3]$. Different number of components $n_p$ were extracted.

### 4.3 Results for fair regression

We analyze the performance of both linear (FLR) and the nonlinear kernel (FKR) formulations. As done in the toy example, we explore the possibility of including or not the sensitive variables **S** in the models. Figure 4 shows the results for different values of $\mu$. Since the original data was collected for a classification problem, we binarized the outputs ($c = 2$), and treat it as a regression problem, afterwards we use max-vote to obtain the predicted class. We analyze two different situations: one where the methods avoid discriminating only by gender and another when the methods avoid discriminating by gender and race simultaneously. Note that in the latter case the sensitive variable is bidimensional. While this situation is quite general, using complicated information measures like mutual information (as proposed in [20]) increases dramatically the complexity of the problem. However, in our case, it is straightforward to deal with multidimensional sensitive variables.

In both cases we observe a similar behavior as in the toy example. Both the linear and kernel classical versions (LR and KRR) obtain relatively good classification error rates, but their dependence with the sensitive variables is relatively high. The use of fair versions open the possibility of decreasing this dependence while yielding similar classification errors. Results are better when using the kernel version FKR, which is capable of learning a model with low classification error rate and virtually independent of the sensitive features. Including the sensitive variables when using our proposed method obtains better results in the kernel case. In the linear case, removing the sensitive features has almost no impact on the results.

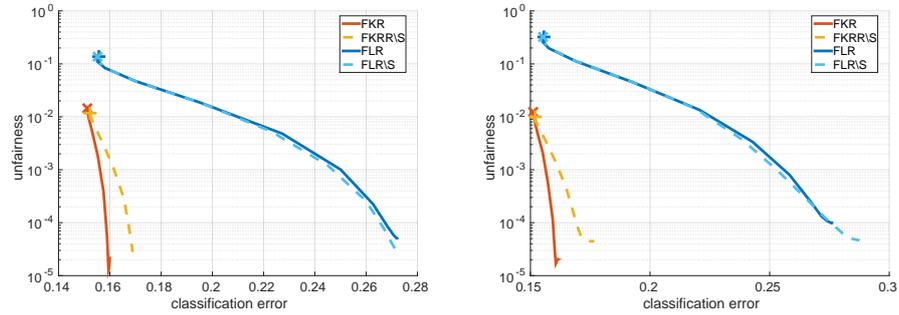

**Fig. 4.** Error in income classification versus (un)fairness of the sensitive variables for the Adult dataset, avoiding discrimination by gender (left), and by both gender and race (right).

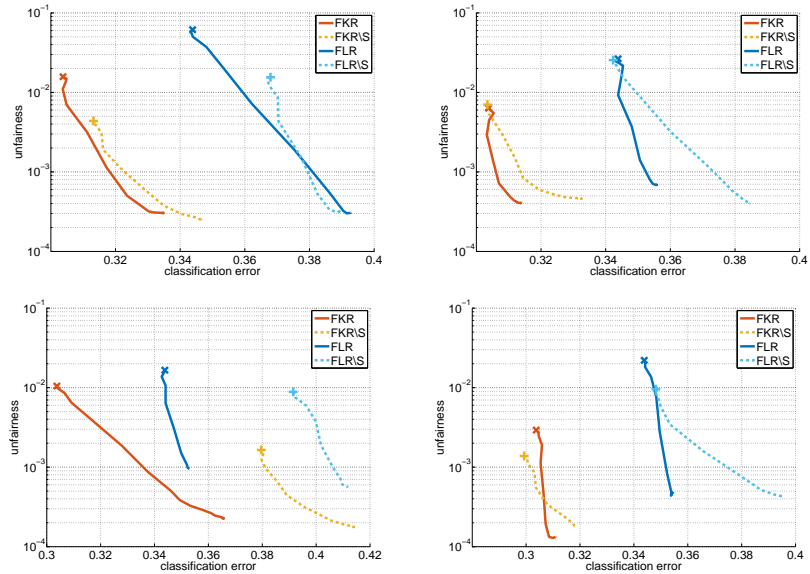

**Fig. 5.** Error in contraceptive usage classification versus (un)fairness of the sensitive variables for the CMC dataset. Top row (left) wife's education, (right) husband's education, and bottom row (left) number of children ever born and (right) media exposure.

When it comes to the second dataset, we performed our experimentation over the sensitive variables: wife's education, husband education, number of children ever born and also media exposure (features 2, 3, 4 and 9 respectively). The experiments were done by considering only one sensitive variable at a time. Figure 5 shows the results for all these protected variables. Several conclusions can be derived: 1) kernel fair regression outperforms the linear counterpart in all the hyperparameter space (both on error and fairness); 2) removing the sensitive feature degrades results, as its information is implicitly conveyed by other included features; and 3) one can achieve arbitrary fairness levels tuning the $\mu$ hyperparameter, at the cost of a moderately increased prediction error (+2-5% increase in classification error).

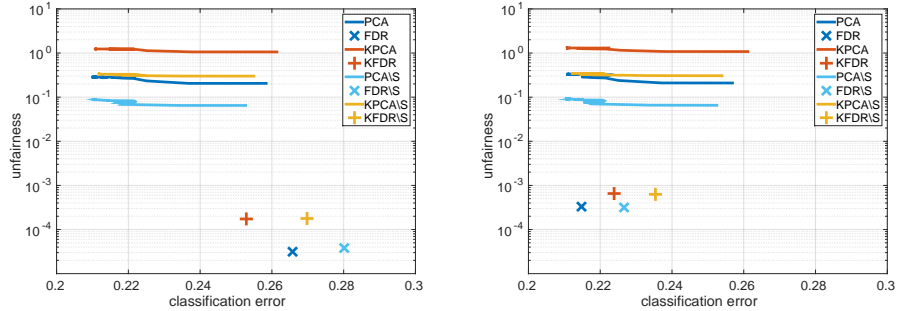

**Fig. 6.** Error rate in income classification versus independence between predictions and the sensitive variables to avoid discrimination by gender (left), and by both gender and race (right). PCA and KPCA has been evaluated for different number of features, $n_p$.

### 4.4 Results for fair dimensionality reduction

We analyzed the performance of the proposed dimensionality reduction in the income prediction dataset. We present results of using a $k$-nn classifier ($k = 1$) after reducing the dimensionality of the data set using different methods. In particular, we analyzed the standard Principal Components Analysis (PCA), Kernel PCA (KPCA), and the proposed fair dimensionality Reduction (FDR) and its kernel counterpart (KFDR). As in the previous experiment, we also analyze the solution with and without the sensitive features as inputs.

Figure 6 shows the solutions of different methods. In the case of PCA and KPCA we show results for different numbers of features, which affects the classification error but minimally the fairness score. In both experiments the best fairness-accuracy trade-offs are given by the FDR and KFDR when using all variables as inputs. In particular, when avoiding the gender discrimination, the proposed framework shows better classification error for the KFDR. When we use as sensitive variables gender and race the differences of the proposals with regard the classical methods are more noticeable since the classification errors are similar but the dependence achieved by the proposals are several orders of magnitude lower.

## 5 Conclusions

We have presented novel fair nonlinear regression and dimensionality reduction methods. We included a term to the cost function based on the Hilbert-Schmidt independence criterion which enforces fairness in the solutions and allows to deal with several sensitive variables simultaneously. We presented the methods in linear fashion and extended them to deal with nonlinear problems by using kernel functions. For both the linear and nonlinear cases, the solution for the regression weights and the basis functions in dimensionality reduction are expressed in closed-form, as they only involve solving matrix inversion or generalized eigenproblems respectively.

Tuning the fairness hyperparameter in regression allows us to input sensitive variables to the regression model while keeping the solution fair. This increases the information that can be used by the model during the prediction rather than just ignoring them. Methods performance were successfully illustrated using both synthetic and real data.

We would like to highlight that introducing kernels (and adopting HSIC) for fairness is not incidental: it allows us to achieve closed-form solutions, to trim fairness-fitness with a single hyperparameter, and to encode prior knowledge in a simple way. Interpretability of the models is obviously an issue and will be explored in the near future. While the framework aims to deal with 'population fairness', not with 'individuals' fairness', this refinement can be easily included in our kernel formulations by defining an individual/group diagonal matrix $\mathbf{F}$ and replacing $\mathbf{X}$ with $\mathbf{XF}$ ($\mathbf{I}$ with $\mathbf{F}^{-1}$ for the kernel formulations). As a future work, we also aim to include kernel conditional independence tests. The proposed framework could be easily extended to other machine learning algorithms, from neural networks to Gaussian processes.